\documentclass[compress,preprint,12pt]{elsarticle}

\usepackage{amssymb}
\usepackage{amsmath}
\usepackage{graphicx}
\usepackage{caption}
\graphicspath{{Images/}}
\usepackage{booktabs}
\usepackage{multirow}
\usepackage{subcaption}
\usepackage{tabularx}
\usepackage{amsmath}
\usepackage{amssymb}
\usepackage{setspace}
\usepackage{graphbox}
\usepackage{adjustbox}
\usepackage{color}
\usepackage{algorithmic}
\usepackage{algorithm}
\usepackage{longtable} 
\usepackage[colorlinks,
            linkcolor=blue,
            anchorcolor=blue,
            citecolor=blue,
            unicode=true]{hyperref}
\usepackage{cleveref}

\floatname{algorithm}{Algorithm}

\crefname{algorithm}{Algorithm}{Algorithms}

\journal{Information Fusion}

\begin{document}

\begin{frontmatter}



\title{Bidirectional Image-Event Guided Fusion Framework for Low-Light Image Enhancement}


\author[author1]{Zhanwen Liu}\corref{1}
\author[author1]{Huanna Song}
\author[author1]{Yang Wang}
\author[author1]{Nan Yang}
\author[author2]{Weiping Ding}
\author[author1]{Yisheng An}
% \cortext[1]{Corresponding author. E-mail address: zwliu@chd.edu.cn}
\cortext[1]{Corresponding author.\\
\hangindent=18pt E-mail addresses: zwliu@chd.edu.cn (Z. Liu), 2024124083@chd.edu.cn (H. Song), ywang120@chd.edu.cn (Y. Wang), 2022024001@chd.edu.cn (N. Yang), ding.wp@ntu.edu.cn (W. Ding), aysm@chd.edu.cn (Y. An)}

\affiliation[author1]{organization={School of Information Engineering, Chang'an University},%Department and Organization
            % addressline={Xi'an}, 
            city={Xi'an},
            postcode={710018}, 
            state={Shaanxi},
            country={China}}
\affiliation[author2]{organization={School of Artificial Intelligence and Computer Science, Nantong University},%Department and Organization
            % addressline={Nantong}, 
            city={Nantong},
            postcode={226019}, 
            state={Jiangsu},
            country={China}}

%% Abstract
\begin{abstract}
Under extreme low-light conditions, frame-based cameras suffer from severe detail loss due to limited dynamic range. 
Recent studies have introduced event cameras for event-guided low-light image enhancement. 
However, existing approaches often overlook the flickering artifacts and structural discontinuities caused by dynamic illumination changes and event sparsity.
To address these challenges, we propose BiLIE, a Bidirectional image-event guided fusion framework for Low-Light Image Enhancement, which achieves mutual guidance and complementary enhancement between the two modalities. 
First, to highlight edge details, we develop a Dynamic Adaptive Filtering Enhancement (DAFE) module that performs adaptive high-pass filtering on event representations to suppress flickering artifacts and preserve high-frequency information under varying illumination.
Subsequently, we design a Bidirectional Guided Awareness Fusion (BGAF) mechanism, which achieves breakpoint-aware restoration from images to events and structure-aware enhancement from events to images through a two-stage attention mechanism, establishing cross-modal consistency, thereby producing a clear, smooth, and structurally intact fused representation.
Moreover, recognizing that existing datasets exhibit insufficient ground-truth fidelity and color accuracy, we construct a high-quality low-light image-event dataset (RELIE) via a reliable ground truth refinement scheme. 
Extensive experiments demonstrate that our method outperforms existing approaches on both the RELIE and LIE datasets.
Notably, on RELIE, BiLIE exceeds the state-of-the-art by 0.81dB in PSNR and shows significant advantages in edge restoration, color fidelity, and noise suppression.
\end{abstract}


% \begin{highlights}
% \item A new high-quality dataset with 2,217 aligned low-light image and event pairs
% \item Dynamic adaptive filtering module suppresses flicker noise and enhances event edges
% \item Bidirectional image–event guidance completes structural discontinuities
% % \item We outperform the state-of-the-art by 0.81dB in PSNR on the proposed dataset
% \item Our method outperforms state-of-the-art approaches on both datasets
% \end{highlights}

%% Keywords

\end{frontmatter}



\section{Introduction}
\begin{figure*}[ht]
\centering
\includegraphics[width=\linewidth]{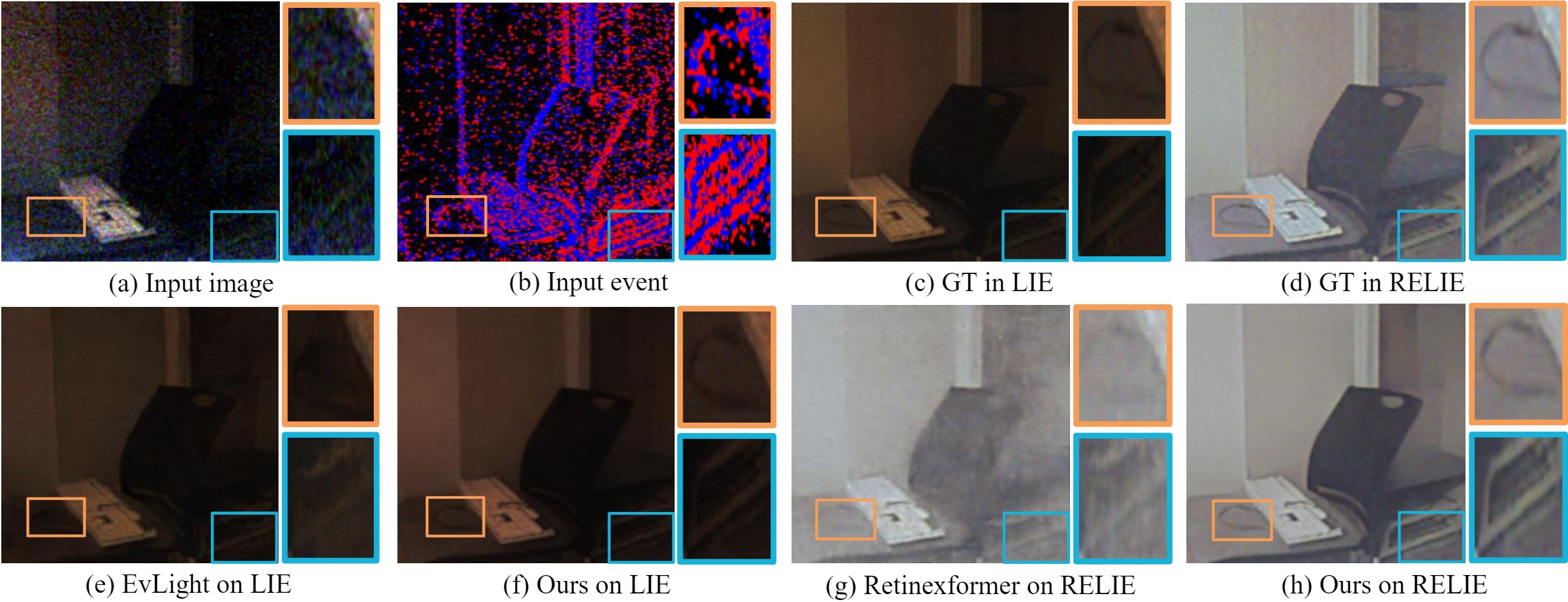}
\caption{The Enhancement results of the frame-based method \cite{cai2023retinexformer}, image-event fusion-based method \cite{liang2024towards} and the proposed method on LIE and our constructed RELIE. Comparing (c) and (d), our constructed dataset exhibits higher image quality. 
The second row demonstrates that our method effectively suppresses noise and artifacts, resulting in higher-quality enhanced images. 
Benefiting from our Dynamic Adaptive Filtering and Bidirectional Guided Awareness fusion design, along with the use of frequency loss, our method effectively suppresses the noise and artifacts introduced during ground truth construction, resulting in smoother and more natural visual outputs.}
\label{Figure 1.} 
\end{figure*}

High-quality low-light image enhancement is a crucial prerequisite for visual perception in scenarios such as intelligent surveillance and autonomous driving \cite{liu2024boosting}.
In recent years, with the rapid development of deep learning, frame-based low-light image enhancement methods \cite{yang2023lightingnet,cai2023retinexformer,li2024light,lin2022low,xu2022snr} have made significant progress, which improve image quality by addressing critical issues such as noise, artifacts, and color distortion. 
However, under extreme low-light conditions, traditional frame-based cameras face the challenge of detail loss, severely limiting the performance of existing methods and making it difficult to reconstruct clear natural-light images, as shown in \autoref{Figure 1.}(g). 

To overcome this bottleneck, researchers have begun to incorporate event cameras into low-light image enhancement \cite{bi2023non,liu2023low,jiang2023event,wang2023event,liang2024towards,liang2023coherent}. 
Event cameras, with unique advantages of high dynamic range and microsecond-level temporal resolution, provide a promising solution for low-light image enhancement. 
By fusing image with event data, these approaches have achieved significant performance improvements. 
However, existing methods primarily adopt an event-guided strategy to compensate for missing structural information in images, which faces three critical challenges. 
First, the quality of current low-light image–event datasets is generally low. Synthetic data sets \cite{liu2023low,wang2023event,liang2024towards,liang2023coherent} struggle to faithfully reproduce complex illumination variations, while real datasets captured with the DAVIS346 event camera \cite{bi2023non,jiang2023event,liang2024towards} are restricted by limited resolution and signal-to-noise ratio.
The reference images exhibit pronounced noise, low contrast, and severe color cast, as shown in \autoref{Figure 1.}(c), which substantially restricts the model performance ceiling. 
Second, the differential sensitivity of event data to intensity variations makes it susceptible to global illumination fluctuations. 
When ambient light changes abruptly, the event stream exhibits a surge of flickering artifacts, as illustrated in the feature map $F_{event}$ in \autoref{Figure4-DAFE.}(c). 
However, most existing studies overlook this low-frequency disturbance characteristics of event data under dynamic illumination conditions.
Third, due to the asynchronous and independent imaging principle of event cameras, the resulting data exhibit spatial sparsity. This characteristic leads to incomplete structural information under event guidance, which is easily affected by discontinuities and results in structural breakages in the fused image, as shown in \autoref{Figure5-BGAF.}(a). 

To address these issues, we systematically improve the public LIE dataset \cite{jiang2023event} by enhancing its reference images with the state-of-the-art unsupervised enhancement methods of the past five years \cite{shi2024zero,ma2022toward,yang2023implicit,li2021learning,fu2023learning}. 
Leveraging a multi-dimensional evaluation protocol that integrates subjective user selection, no-reference objective metrics, and large-model-assisted scoring, we construct a new dataset, RELIE, characterized by high fidelity and superior visual quality. 
As illustrated in \autoref{Figure 1.}(d), RELIE exhibits a pronounced visual advantage over the original dataset.

Leveraging the newly constructed dataset RELIE, we propose an innovative Bidirectional image–event guided Low-light Image Enhancement framework (BiLIE), which comprises two core components: Dynamic Adaptive Filtering Enhancement (DAFE) module and Bidirectional Guided Awareness Fusion (BGAF) module. 
Specifically, the DAFE module employs a dual-branch architecture coupled with a dynamic gating fusion mechanism to perform adaptive high-pass filtering, effectively suppressing global artifacts in the event representation while preserving critical edges and fine details, thereby ensuring the extraction of a more discriminative event representation. 
To address structural breakpoints in the event stream, the BGAF module introduces a two-stage guided attention mechanism that enables bidirectional enhancement between images and events. 
In the first stage, the spatial continuity and structural consistency of the image are extracted and used to guide breakpoint completion and color restoration for events (\autoref{Figure5-BGAF.}c).
In the second stage, the high sensitivity of events to luminance changes reinforces global structure and dynamic details in the image (\autoref{Figure5-BGAF.}d), yielding a fused representation that is clearer, more complete, and smoother. 
Additionally, we incorporate frequency loss and color consistency loss to further reduce noise and artifacts while maintaining color fidelity. Experimental results demonstrate that our method achieves state-of-the-art performance on both the LIE and RELIE datasets, particularly excelling in noise suppression and smoothness, as illustrated in \autoref{Figure 1.}(f) and (h).

In summary, our contributions include the following four aspects:

(1) To overcome the poor visual quality of existing datasets, we construct a high-quality low-light enhancement dataset (RELIE), containing 2,217 strictly aligned pairs of low-light images and events, along with corresponding reference images.

(2) To address the low-frequency disturbance characteristics exhibited by event data under dynamic illumination, we design a Dynamic Adaptive Filtering Enhancement (DAFE) module that effectively suppresses flickering artifacts while preserving high-frequency event information.

(3) We further propose a Bidirectional guided Low-Light Image Enhancement framework (BiLIE), incorporating a Bidirectional Guided Awareness Fusion (BGAF) mechanism to achieve structural completion and detail enhancement. 
In addition, frequency loss and color consistency loss are introduced to generate outputs with reduced noise and improved color fidelity, effectively addressing structural discontinuities caused by sparse events and color shifts introduced by camera limitations.

(4) Extensive experiments demonstrate that BiLIE achieves state-of-the-art performance on both the LIE and RELIE datasets, enabling high-quality reconstruction under extremely low-light conditions. 
Notably, it outperforms existing methods on the RELIE dataset by a PSNR margin of 0.81dB.

The remainder of this paper is organized as follows: \autoref{sec:section 2} reviews event-based low-light image enhancement datasets and existing studies on low-light enhancement algorithms.
\autoref{sec:section 3} introduces the construction process and multidimensional evaluation scheme of the proposed high-quality low-light image–event dataset (RELIE).
\autoref{sec:section 4} elaborates on the proposed Bidirectional Image–Event Fusion Framework (BiLIE), including the Dynamic Adaptive Filtering Enhancement (DAFE) module, the Bidirectional Guided Awareness Fusion (BGAF) mechanism, and the design of the loss functions adopted in the network.
\autoref{sec:section 5} presents extensive experimental results and ablation analyses on the RELIE and LIE datasets, followed by both quantitative and qualitative discussions on model performance.
Finally, \autoref{sec:section 6} concludes the paper and discusses future research directions.

\section{Related Work}
\label{sec:section 2}
\subsection{Event-based Low-light Enhancement Datasets}
Early datasets primarily served traditional image enhancement methods \cite{liu2022dynamic,lee2013adaptive}, which imposed relatively low requirements on data scale and quality. 
With the advancement of deep learning, learning-based methods \cite{yang2023lightingnet,cai2023retinexformer,li2024light,lin2022low,xu2022snr} have become increasingly dependent on the quality of training data. 
In this paper, we focus on low-light image-event datasets.
To date, EvLowLight \cite{liang2023coherent}, EvLight \cite{liang2024towards}, Liu et al. \cite{liu2023low}, and Wang et al. \cite{wang2023event} have provided synthetic events for four image datasets: Davis2017 \cite{zhang2021learning}, SDSD \cite{wang2021seeing}, Vimeo90k \cite{xue2019video}, and LOL \cite{wei2018deep}.  
Although synthetic approaches are easy to implement, they suffer from significant domain gaps and have limited generalization to real-world scenarios.
In recent years, researchers have employed the DAVIS346 camera to construct real-world datasets. 
For example, DAVIS-NUIUIED \cite{bi2023non} targets image enhancement under non-uniform illumination; LIE \cite{jiang2023event} provides paired low-light/normal-light images and low-light events for static scenes; and SDE \cite{liang2024towards} captures paired image-event sequences using a robotic arm equipped with DAVIS346. 
However, the inherent low resolution and pronounced color shift of this camera yield poor-quality normal-light images, which severely constrain the performance ceiling of learning-based methods.
In contrast, our constructed RELIE dataset systematically improves the realism and reliability of reference images through unsupervised enhancement and a multi-dimensional evaluation mechanism.

\subsection{Low-light Image Enhancement Methods}
\textbf{Frame-Based Low-Light Image Enhancement.} Traditional image enhancement methods include histogram equalization \cite{liu2022dynamic} and Retinex-based approaches \cite{lee2013adaptive}. 
Histogram equalization enhances local contrast to reveal image details but often amplifies noise and introduces artifacts. 
Retinex-based methods model illumination and reflectance components under the assumption of color constancy, aiming to improve image quality.
However, they still suffer from detail loss under complex lighting conditions.
With the development of deep learning, convolutional neural networks (CNNs) \cite{li2024light,lin2022low} have been widely adopted to enhance low-light images by learning complex structures and semantic features from large-scale datasets. 
Nonetheless, due to their limited receptive fields, these methods struggle to capture long-range dependencies effectively.
Recently, Transformer leverages attention mechanisms to model global features, has demonstrated strong performance in image enhancement and restoration tasks \cite{yang2023lightingnet,cai2023retinexformer,xu2022snr,huang2024df3net,shang2025lgt,fan2025color}. However, it is important to note that frame-based cameras, constrained by limited dynamic range and temporal resolution, still suffer from detail loss under extremely low-light conditions. 
As a result, the performance of the above methods is significantly limited in such challenging scenarios.

\textbf{Event-Based Low-Light Image Enhancement.} Event cameras are bio-inspired vision sensors that asynchronously capture dynamic changes in brightness \cite{rebecq2019events}. 
With their high dynamic range and microsecond-level temporal resolution, event cameras can detect rapid luminance changes and provide accurate motion and edge information, even in low-light environments. 
Such characteristics have demonstrated outstanding performance in autonomous driving applications \cite{liu2024enhancing, shi2025fusion,liu2025beyond}.
Some studies \cite{rebecq2019high,wang2019event,cadena2021spade,zou2021learning,liu2024seeing,zhang2020learning} have explored the possibility of reconstructing clear images from event. 
For example, Liu et al. \cite{liu2024seeing} proposes a nighttime event-based reconstruction network that generates visually pleasing high dynamic range images under non-uniform illumination and high-noise conditions. However, purely event-based methods often lack sufficient color information.
In recent years, researchers have focused on event-guided fusion methods to improve the visual quality of low-light images. 
Liang et al. \cite{liang2023coherent} establishes spatio-temporal consistency across different modalities and resolutions by modeling correlations across space and time. 
ELIE \cite{jiang2023event} fuses events and images by leveraging residual information between the two modalities. 
Wang et al. \cite{wang2023event} proposes a dual-branch event-guided attention fusion network, consisting of a gradient reconstruction branch and an image enhancement branch. 
EvLight \cite{liang2024towards} introduces snr-guided feature selection, adaptively extracting image features from high-SNR regions and event features from low-SNR regions.

% Despite the promising results of these approaches, they generally overlook two key issues: global flickering artifacts in event representations under dynamic illumination and structural discontinuities arising from the spatial sparsity of event data. 
% In contrast, our Bidirectional image-event guided Low-light Image Enhancement framework (BiLIE) leverages DAFE and BGAF modules to suppress global artifacts and reconstruct structurally complete representations while preserving high-frequency structural details.
As discussed, although extensive research efforts have been devoted to image enhancement methods, it is worth noting that existing algorithms still suffer from the following limitations and challenges.
1) Event representations are highly sensitive to brightness changes, and current methods often fail to suppress global flickering artifacts when illumination varies rapidly.
2) The spatially sparse and uneven distribution of events causes incomplete or fragmented edge structures, leading to structural inconsistencies in the reconstructed results.
3) Most existing methods rely solely on a one-way event-to-image guidance strategy, failing to fully exploit the complementary advantages of dynamic information from the event modality and spatial structural information from the image modality.
In contrast,our proposed Bidirectional Image–Event Guided Low-Light Image Enhancement (BiLIE) framework introduces a Dynamic Adaptive Filtering Enhancement (DAFE) module to adaptively suppress flickering artifacts and preserve high-frequency details, and a Bidirectional Guided Awareness Fusion (BGAF) mechanism to establish cross-modal consistency and achieve structure-aware enhancement and completion.

\section{RELIE Dataset}
\label{sec:section 3}
\begin{figure*}[ht]
    \centering
    % 上边：图像
    \begin{subfigure}[t]{\linewidth}
        \centering
        \includegraphics[width=\linewidth]{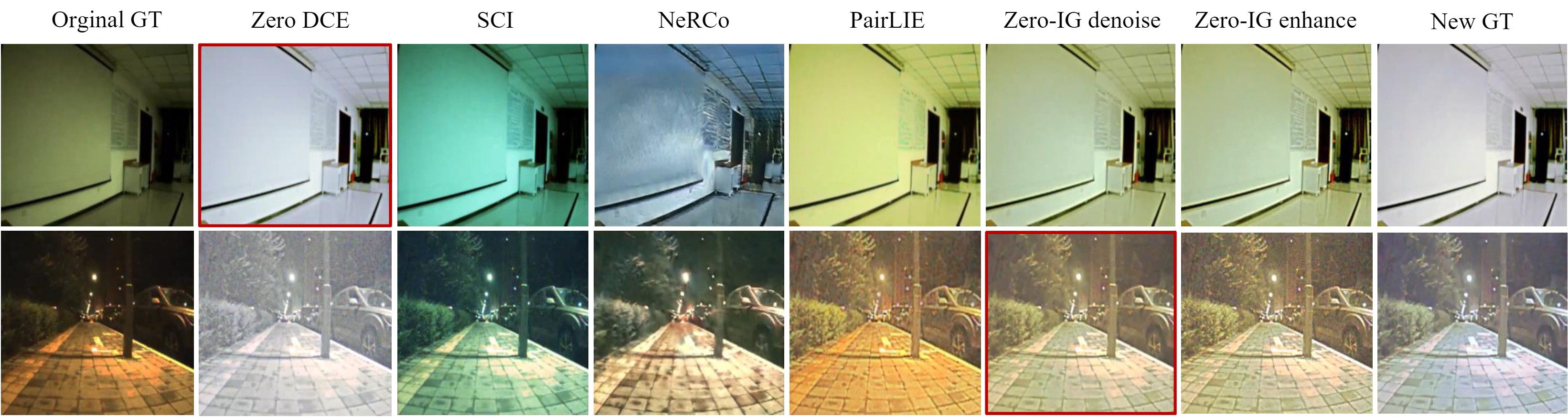}
        \caption*{(a) Results of different unsupervised enhancement methods}
    \end{subfigure}
    % 左边：图像
    \begin{minipage}{0.35\linewidth}
        \begin{flushleft}
            % \centering
            % \caption*{(b) Percentage of different methods}
            \includegraphics[width=\linewidth, height=3.8cm]{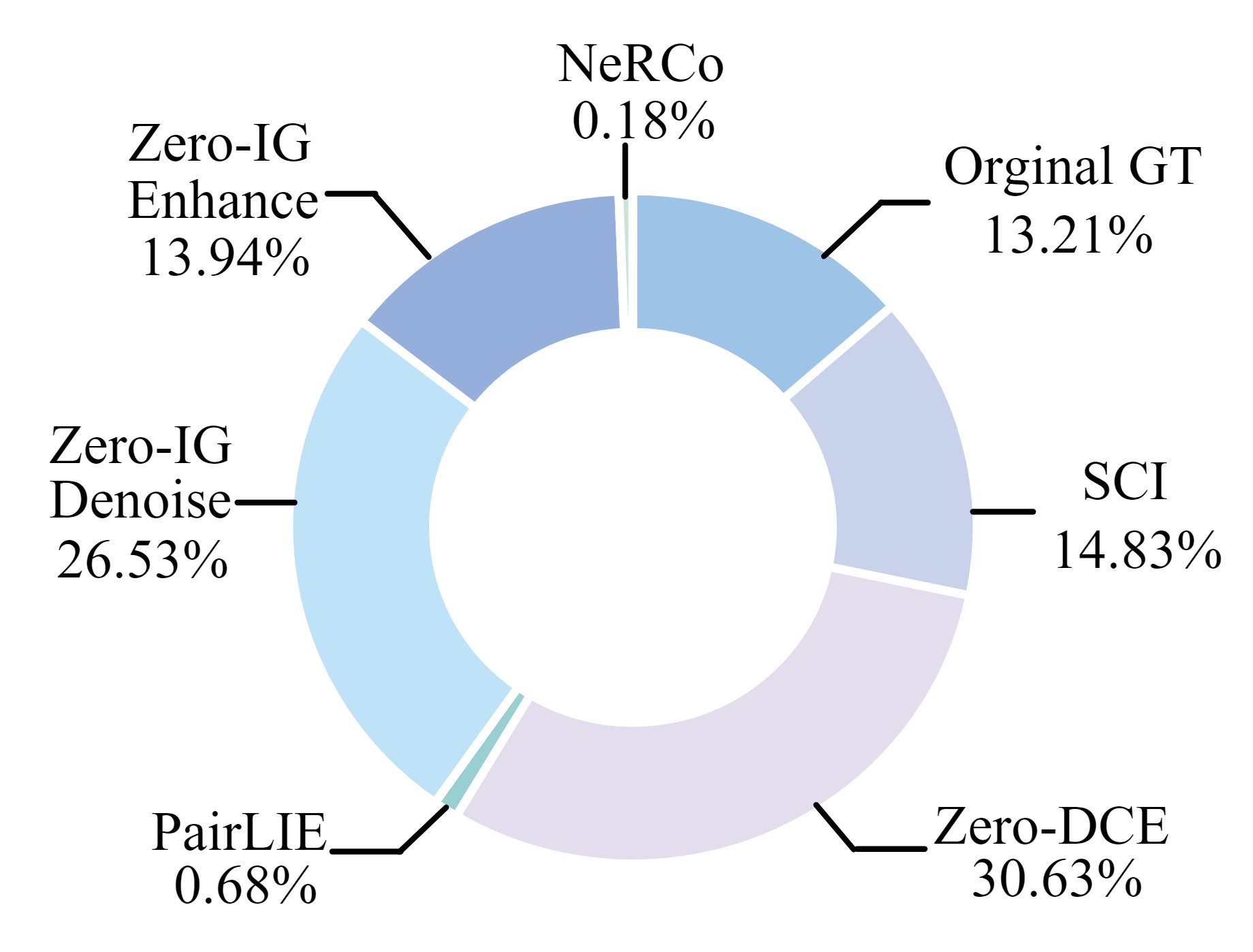}
            \caption*{(b) Method-wise percentage}
        \end{flushleft}
    \end{minipage}
    % 右边：表格
    \begin{minipage}{0.64\textwidth}
        \footnotesize
        \centering
        \renewcommand\arraystretch{0.9}
        \setlength{\tabcolsep}{2.5pt}
        % \caption*{(c) Ground Truth Quality Evaluation Table}
        \begin{adjustbox}{width=\linewidth}
        \begin{tabularx}{\linewidth}{cccc}
            \toprule
            Evaluation Method & Source & RELIE & LIE \\
            \midrule
            \adjustbox{valign = m}{\includegraphics[width = 1.1cm]{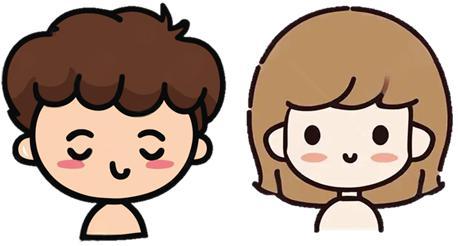}}~Subjectivity
            & User Study & 86.29\%  & 13.71\%  \\
            \midrule
            \multirow{2}{*}{
                \adjustbox{valign = m}{\includegraphics[width = 0.9cm, height=0.8cm]{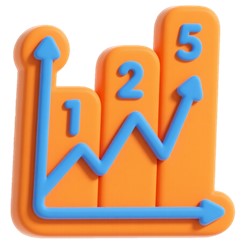}}~Objectivity
            }
            & PIQE $\downarrow$ & 7.7 & 10.5 \\
            & BRISQUE $\uparrow$ & 1.93 & 1.66 \\
            \midrule
            \multirow{3}{*}{
                \adjustbox{valign = m}{\includegraphics[width = 0.9cm]{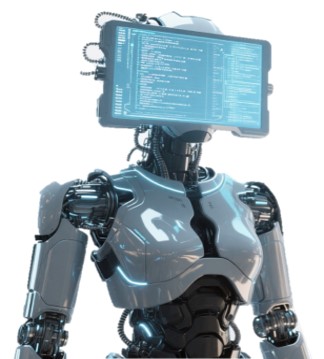}}~Large Model
            }
            & Tencent HunYuan & 95.56\% & 4.44\% \\
            & Qwen2.5-VL-72B & 99.01\% & 0.99\% \\
            & Gemini-2.5-Pro & 98.03\% & 1.97\% \\
            \bottomrule
        \end{tabularx}
        \end{adjustbox}
        \caption*{(c) Ground truth quality evaluation}
    \end{minipage}
    \hfill
    \caption{Comparison of Ground Truth Quality between RELIE and LIE Datasets: (a) From left to right: the ground truth of the LIE dataset \cite{jiang2023event}, results of Zero-DCE \cite{li2021learning}, SCI \cite{ma2022toward}, NeRCo \cite{yang2023implicit}, PairLIE \cite{fu2023learning}, Zero-IG \cite{shi2024zero}, and the ground truth of the new dataset RELIE. The \textcolor{red}{red} box highlights the candidate results, which are further white-balanced to obtain the final ground truth. (b) The ground truth in RELIE comes from the proportion of different enhancement methods. (c) Evaluation results from subjective, objective, and large model.}
    \label{Figure2-dataset.}
\end{figure*}

After systematically reviewing previous works, we observe that the ground truth images provided by existing datasets suffer from limited contrast, color distortion, and detail loss due to the hardware limitations of the DAVIS346 camera, as shown in the first column of \autoref{Figure2-dataset.}(a). These deficiencies severely hinder further improvements in model performance in terms of both visual quality and perceptual realism.
To address this issue, we construct a new dataset named RELIE based on the LIE dataset \cite{jiang2023event}. 
We apply five state-of-the-art unsupervised low-light image enhancement methods to enhance the 2,217 reference images in the LIE dataset. 
These methods include Zero-Reference Deep Curve Estimation (Zero-DCE \cite{li2021learning} ), Self-Calibrated Illumination Learning (SCI \cite{ma2022toward}), NeRCo \cite{yang2023implicit} based on implicit neural representation, and two Retinex-based methods: PairLIE \cite{fu2023learning} and Zero-IG \cite{shi2024zero}. 
The source code for all methods is provided by their respective authors.
As a result, we generate six enhanced versions for each reference image, including two versions from Zero-IG (before and after denoising), leading to a total of 6 × 2,217 enhanced images.

For ground truth selection, we design a human-in-the-loop collaborative filtering process. 
We invite 11 volunteers with a background in image processing to evaluate each group of seven images (including the original reference and six enhanced results), as illustrated in \autoref{Figure2-dataset.}(a). 
The volunteers select the image that best approximates the real scene while exhibiting the highest visual quality, considering four criteria: noise level, contrast, color naturalness, and artifact presence.
After all participants complete their evaluations, we collect and aggregate the votes. 
For each group, we highlight the most frequently selected image with a red bounding box, which is then chosen as the candidate ground truth. 
The percentage of images selected as optimal for each method is shown in \autoref{Figure2-dataset.}(b), with Zero-DCE and Zero-IG denoise ranking first and second, respectively.

Finally, to further improve the color consistency and realism of the candidate ground truth images, we apply a white balance correction based on the gray-world assumption. 
As shown in the last column of \autoref{Figure2-dataset.}(a), the corrected images exhibit more natural and realistic colors, making them more visually aligned with human perception and more suitable as ground truth in our newly constructed dataset.
Specifically, we use the mean luminance of the green channel as a reference to compute gain factors for the red and blue channels, which are then adjusted accordingly. 
The three channels are subsequently merged to obtain the final corrected image. 
We select the green channel as the reference because it is generally less sensitive to illumination variations than the red and blue channels, and its average intensity tends to be more stable \cite{mccartney1976optics}. 
This stability makes it a reliable indicator of the overall brightness and color tone of an image, thereby improving the accuracy of the white balance correction.

To evaluate the improvements in image quality achieved by RELIE over LIE, we introduce a multi-dimensional evaluation mechanism, as illustrated in \autoref{Figure2-dataset.}(c). 
Based on user study results, only 13.21$\%$ of the selected ground truth images originate from the original LIE dataset, while 86.79$\%$ are selected from the unsupervised enhancement results.
We compare the ground truth images of both RELIE and LIE with two no-reference objective metrics: Perception-based Image Quality Evaluator (PIQE) \cite{venkatanath2015blind} and Blind/Referenceless Image Spatial Quality Evaluator (BRISQUE) \cite{mittal2012no}.
The results indicate that RELIE consistently achieves superior performance across both metrics.
Furthermore, we utilize three advanced vision language models, including Tencent HunYuan, Qwen2.5-VL-72B, and Gemini-2.5-Pro, to perform AI-assisted quality assessment. 
The results show that all three models prefer the ground truth images from RELIE over those from LIE in more than 95$\%$, further confirming the reliability and visual quality of our constructed dataset.

\section{Proposed Method}
\label{sec:section 4}
\begin{figure*}[ht]
\centering 
\includegraphics[width=\linewidth]{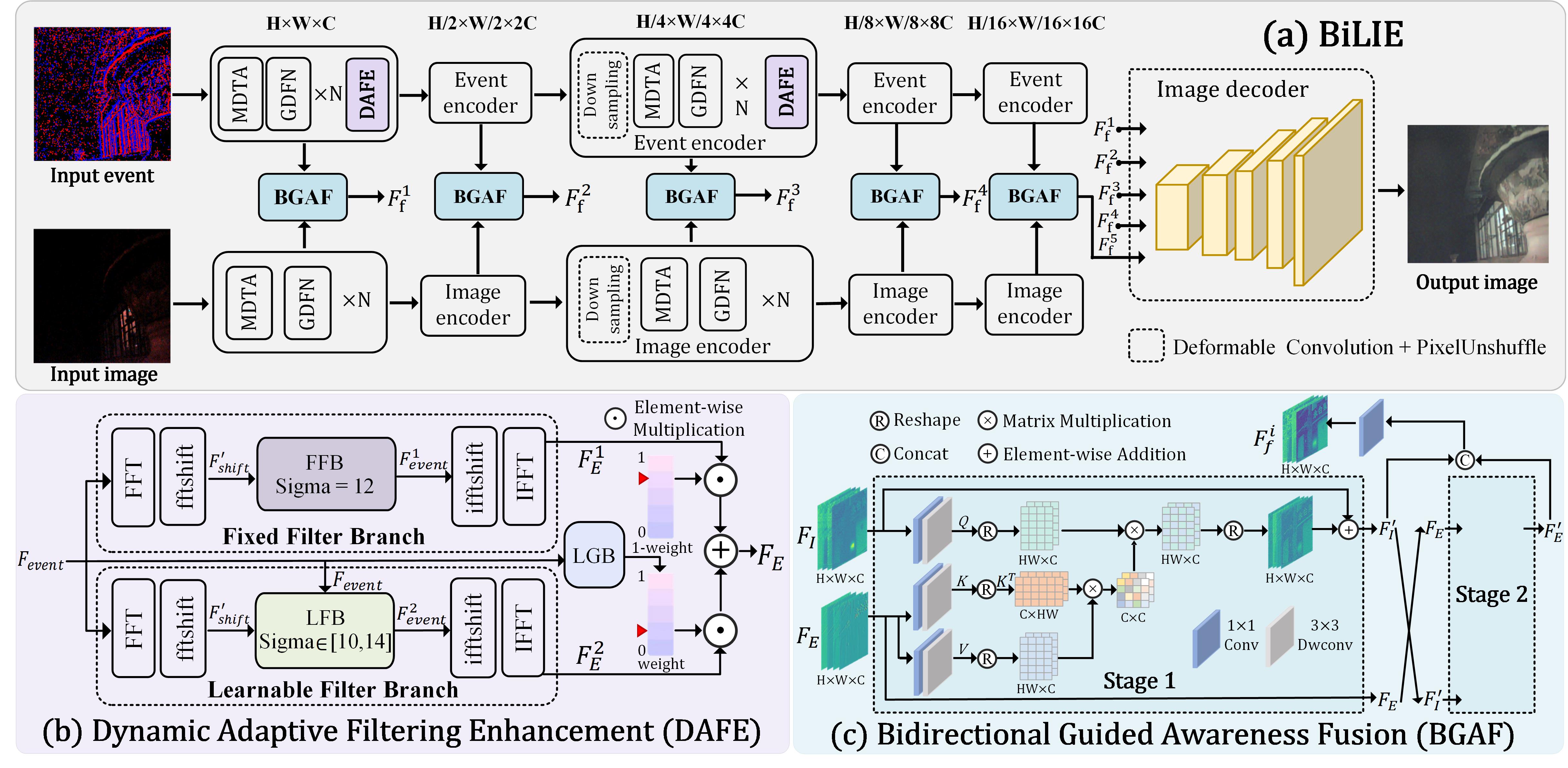} 
\caption{The network architecture of our proposed method. BiLIE adopts a dual-branch encoder-decoder structure, consisting of two fundamental units: Dynamic Adaptive Filtering Enhancement (DAFE) and Bidirectional Guided Awareness Fusion (BCAF). These units work together to generate high-quality, smooth images with clear contours.}
\label{Figure3-BiLIE-network.} 
\end{figure*}

\subsection{Method Overview}
As illustrated in \autoref{Figure3-BiLIE-network.}, BiLIE takes a low-light image $F \in R^{3\times H\times W}$ and the corresponding event tensor $E \in R^{5\times H\times W}$ as input. 
The event tensor is constructed using the voxel grid representation proposed by Rebecq et al. \cite{rebecq2019high}, where the polarity of each event is distributed to the two adjacent temporal bins closest to its timestamp.
The temporal binning strategy follows the configuration of ELIE \cite{jiang2023event}.
Modality-specific features are extracted using two separate encoders built on the Restormer architecture \cite{zamir2022restormer}, where each encoder is composed of Multi-Dconv Head Transposed Attention (MDTA) and Gated Dconv Feed-Forward Network (GDFN) modules to capture modality-specific representations for image and event streams, respectively.
To alleviate flickering artifacts caused by dynamic illumination changes, the event features are further refined through the Dynamic Adaptive Filtering Enhancement (DAFE) module, as shown in \autoref{Figure3-BiLIE-network.}(b).
Subsequently, these features are fused with the image branch via the Bidirectional Guided Awareness Fusion (BGAF) module, as shown in \autoref{Figure3-BiLIE-network.}(c), enabling cross-modal interaction that mitigates structural discontinuities resulting from event sparsity and improves the spatial continuity and structural completeness of the fused representation.
Finally, under the joint supervision of four loss functions, BiLIE reconstructs high-quality images with clear details, natural appearance, and faithful color without noticeable artifacts.
To make the proposed pipeline clearer, we summarize the overall procedure in \autoref{alg:BiLIE}.

\begin{algorithm}
    \caption{Overall Pipeline of the BiLIE Framework}
    \label{alg:BiLIE}
    \renewcommand{\algorithmicrequire}{\textbf{Input:}}
    \renewcommand{\algorithmicensure}{\textbf{Output:}}
    
    \begin{algorithmic}[1]
        \REQUIRE Low-light image $F$, event tensor $E$  
        \ENSURE Enhanced image $I_{output}$    
        \FOR{training epoch = 1 to 200} 
            \STATE Preprocess F and E to obtain input features;
            \STATE Extract multi-scale features using the image encoder and event encoder;
            \STATE Apply Gaussian high-pass filtering to the event feature $F_{event}$ according to \autoref{eq1}, \autoref{eq2} and \autoref{eq3} to obtain the fixed filter branch output $F_{E}^{1}$;
            \STATE Compute adaptive parameter $\sigma_{2}$ for each sample using the learnable filter block defined in \autoref{eq4};
            \STATE Perform adaptive filtering according to \autoref{eq2} and \autoref{eq3} to obtain the adaptive filter branch output $F_{E}^{2}$;
            \STATE Calculate dynamic gating weight $weight$ using \autoref{eq5}, and perform dynamic adaptive fusion to produce the DAFE module output $F_{E}$;
            \FOR{each scale $i$}
                \STATE \textbf{Stage 1: Image-guided event enhancement}
                \STATE Update the image feature $F_{I}^{'}$ using \autoref{eq6};
                \STATE \textbf{Stage 2: Event-guided image enhancement}
                \STATE Update the event feature $F_{E}^{'}$ using \autoref{eq7};
                \STATE Concatenate $F_{I}^{'}$ and $F_{E}^{'}$ along the channel dimension according to \autoref{eq8} to obtain the fused feature $F_{f}^{i}$;
            \ENDFOR
            \STATE Upsample and decode all fused multi-scale features $F_{f}^{i}$ to generate the enhanced image $I_{output}$;
            \STATE Compute the hybrid loss function according to \autoref{eq9};
            \STATE Update model parameters based on the loss results;
            % \IF {the model performance does not improve for 30 consecutive epochs;}
            %     \STATE \textbf{break;}
            % \ENDIF
        \ENDFOR
    \end{algorithmic}
\end{algorithm}

\subsection{Dynamic Adaptive Filtering Enhancement}
Under complex and dynamic lighting conditions, event cameras are prone to interference from flicker effects, which become particularly pronounced in scenarios such as streetlights, vehicle headlights, or indoor illumination at night. 
For example, LIE \cite{jiang2023event} triggers indoor event streams by switching lights on and off. 
Such global brightness changes introduce substantial flickering artifacts  into the event representation space, severely degrading perceptual quality.
However, in the event modality, the information we truly care about lies in high-frequency structural details such as edges and textures, which capture the most critical structural variations and perceptual cues. 
Therefore, we design a dynamic adaptive high-pass filtering mechanism to enhance the high-frequency structural information in event features while effectively suppressing low-frequency artifacts interference.
As illustrated in \autoref{Figure3-BiLIE-network.}(b), the module first transforms the event representation into the frequency domain and constructs a dual-branch structure. 
One branch incorporates a fixed prior to provide stable filtering guidance, while the other employs a learnable adjustment branch to enhance adaptability. 
A dynamic gating mechanism then adaptively fuses the outputs of the two branches, achieving a robust and flexible balance between stability and adaptability.
Algorithm 1 shows the implementation process of the DAFE module.

\textbf{Fixed Filter Branch.} First, we perform a two-dimensional discrete Fourier transform on the input event features $F_{event}$, transferring it from the spatial domain to the frequency domain:
\begin{equation}
  F_{event}^{'} \left ( u,v \right ) =\sum_{x=0}^{M-1} \sum_{y=0}^{N-1} F_{event}\left ( x,y \right ) e^{-j2\pi \left ( \frac{ux}{M}+  \frac{vy}{N}   \right ) } 
  \label{eq1}
\end{equation}
where $M$ and $N$ are the number of rows and columns of the image, respectively. 
$(x,y)$ denotes the spatial domain coordinates.
$(u,v)$ represents the frequency domain coordinates.
$j=\sqrt{-1 }$.

\begin{figure*}[ht]
\centering 
\includegraphics[width=\linewidth]{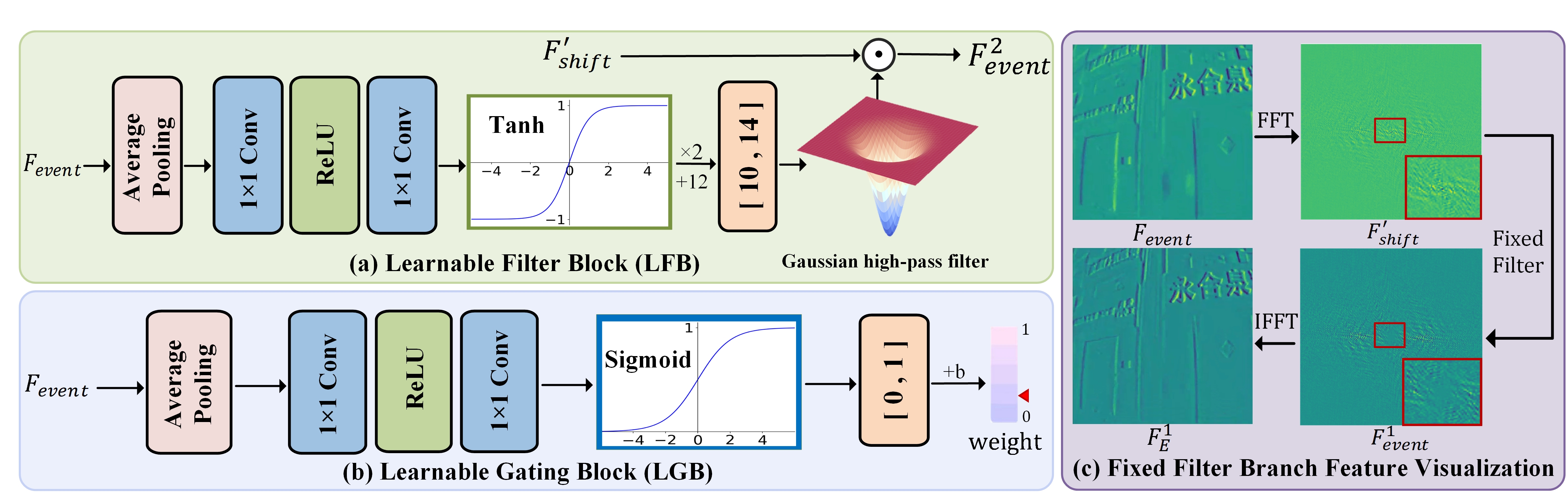} 
\caption{The structures of the Learnable Filter Block (LFB) and the Learnable Gating Block (LGB), as well as the feature visualization results of the Fixed Filter Branch within the Dynamic Adaptive Filtering Enhancement (DAFE) module. $F_{event}$, $F_{shift}^{'}$, $F_{event}^{1}$, and $F_{E}^{1}$ respectively represent the feature map before DAFE processing, after Fourier transform and frequency-domain shifting, after the high-pass filter, and after DAFE processing. Compared to the original event representation $F_{event}$, the filtered result $F_{E}^{1}$ is more natural and sharper.}
\label{Figure4-DAFE.} 
\end{figure*}

The frequency-domain shifted result $F_{shift}^{'} \left ( u,v \right )$ is obtained using the fftshift function.
As shown in the red box of the feature map $F_{shift}^{'}$ in \autoref{Figure4-DAFE.}(c), the prominent bright spot near the center represents low-frequency components, while the gradually fading ripple patterns toward the periphery correspond to edges and high-frequency details.
Next, a Gaussian high-pass filter $High_{1}\left ( u,v \right )$ is constructed and applied by performing element-wise multiplication with the frequency-domain features $F_{shift}^{'} \left ( u,v \right )$. 
This filter exhibits a smooth cutoff behavior, which effectively avoids the ringing artifacts typically caused by ideal high-pass filters:
\begin{equation}
\begin{aligned} 
&High_{1}\left ( u,v \right ) =1-e^{-\frac{\left ( \sqrt{\left ( u-u_{c}  \right ) ^{2}+\left ( v-v_{c}  \right ) ^{2}  }  \right ) ^{2} }{2\sigma_{1} ^{2} } }, \\
&F_{event}^{1} \left ( u,v \right ) =High_{1}\left ( u,v \right ) \odot F_{shift}^{'} \left ( u,v \right ) .
  \label{eq2}
\end{aligned}
\end{equation}
where $u_{c}$ and $v_{c}$ are the frequency domain centers.
$\sqrt{\left ( u-u_{c}  \right ) ^{2} +\left (v-v_{c}   \right ) ^{2}  }$ represents the distance from the frequency domain coordinates $\left ( u,v \right )$ to the frequency domain center $\left (u _{c} ,v_{c}  \right )$.
$\sigma_{1}$ is the parameter that controls the bandwidth of the filter. 
In our experiments, the Fixed Filter Branch uses a standard deviation of 12, which yields optimal filtering performance, as shown in \autoref{table 3}.

After filtering, as shown in \autoref{Figure4-DAFE.}(c), the ripple-like patterns in the outer regions of the feature map $F_{event}^{1} \left ( u,v \right )$ become more pronounced and concentrated, indicating that the low-frequency components are effectively suppressed. 
Finally, the filtered and inverse-shifted representation $F_{shift}^{1}\left (  u,v\right )$ is transformed back to the spatial domain using a two-dimensional inverse discrete Fourier transform to produce the output $F_{E}^{1} \left ( x,y \right )$.
Compared to the original event representation $F_{event}$, the filtered result $F_{E}^{1}$ reduces global brightness noise, achieves a more natural and visually coherent appearance, sharper edge contours, and more prominent high-frequency details:
\begin{equation}
  F_{E}^{1} \left ( x,y \right ) =\frac{1}{MN} \sum_{u=0}^{M-1} \sum_{v=0}^{N-1}F_{shift}^{1}\left (  u,v\right )e^{j2\pi \left ( \frac{ux}{M} +\frac{vy}{N}  \right ) }    
  \label{eq3}
\end{equation}

\textbf{Learnable Filter Branch.} Considering the variations in event features across different scenes, lighting conditions, and motion intensities, we introduce a locally learnable mechanism that adaptively adjusts filtering parameters to enhance the model’s perceptual flexibility and adaptability.
As shown in \autoref{Figure4-DAFE.}(a), the structure of the Learnable Filter Block (LFB) consists of a global average pooling layer followed by a lightweight MLP network (implemented as a 1×1 convolution), which predicts adaptive parameters $\sigma_{2}$ for each sample:
\begin{equation}
    \sigma_{2} = 12+  2\cdot tanh\left ( Conv\left ( ReLU\left ( Conv\left ( GAP\left ( F_{event}  \right )  \right )  \right )  \right )  \right )
    \label{eq4}
\end{equation}
where $\sigma_{2}$ denotes the parameter range of the filter that is adaptively adjusted for each sample according to its context.
$\sigma_{2} \in \left [ 10,14 \right ] $.

The predicted parameter $\sigma_{2}$ is used to construct the corresponding high-pass filter $High_{2}\left ( u,v \right )$. 
We perform frequency-domain filtering and inverse transformation according to \autoref{eq2} and \autoref{eq3}, resulting in an adaptively enhanced event representation $F_{E}^{2}$.

\textbf{Learnable Gating Block.} As illustrated in \autoref{Figure4-DAFE.}(b), to achieve a better balance between fixed frequency priors and adaptive adjustment capacity, we introduce a biased dynamic gating mechanism that controls the weight ratio of the outputs from the two branches, yielding the final fused output:
\begin{equation}
\begin{aligned} 
&weight= \sigma \left ( Conv\left ( ReLU\left ( Conv\left (  GAP\left ( F_{event}  \right ) \right )  \right )  \right )  \right )+ b, \\
&F_{E}   = weight\cdot F_{E}^{2} +  \left (  1-weight\right ) \cdot F_{E}^{1} .
    \label{eq5}
\end{aligned}
\end{equation}
where $b$ denotes the bias term that governs the filter parameters, and $weight$ represents the adaptive control parameter produced by the gating network.

\subsection{Bidirectional Guided Awareness Fusion}

\begin{figure*}[ht]
\centering 
\includegraphics[width=\linewidth]{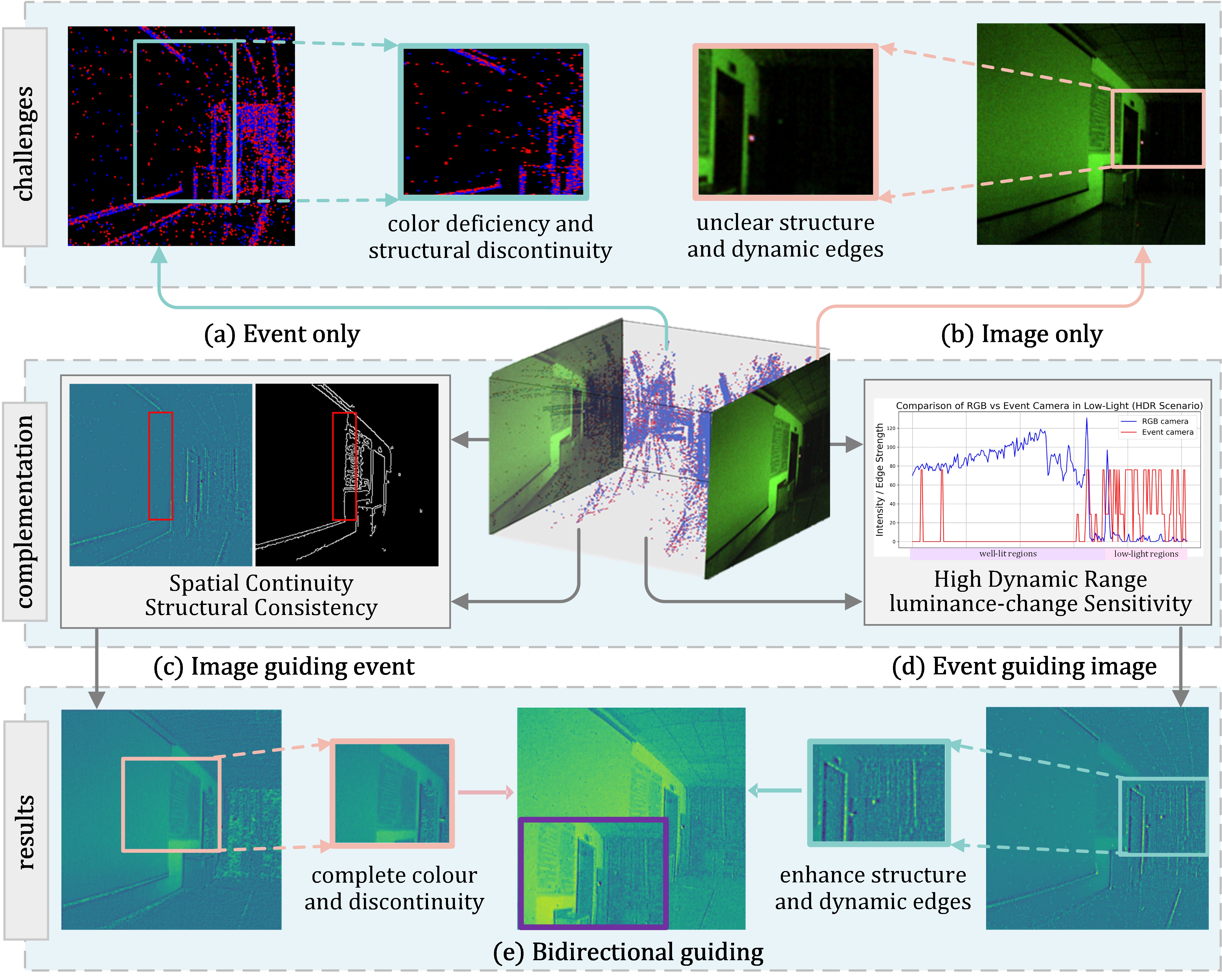} 
\caption{Motivation analysis and feature visualization for bidirectional image–event guidance.} 
\label{Figure5-BGAF.} 
\end{figure*}

Image data and event data exhibit significant differences in visual distributions.
As illustrated in \autoref{Figure5-BGAF.}, the image modality exhibits dense spatial sampling and complete color–texture information, providing global semantic and contextual structure.
However, under extremely low-light conditions, it is constrained by limited dynamic range, often leading to detail loss and noise contamination, as shown in \autoref{Figure5-BGAF.}(b). 
In contrast, event cameras detect pixel-level luminance changes and can accurately capture edge and motion information across a high dynamic range, maintaining strong responsiveness even in extremely dark environments.
As a result, recent studies \cite{bi2023non,liu2023low,jiang2023event,wang2023event,liang2024towards,liang2023coherent} have introduced events as structural guidance to enhance and complete low-light images. 
However, due to the asynchronous imaging principle, event data are inherently spatially sparse and unstructured, which leads to pronounced structural discontinuities, as shown in \autoref{Figure5-BGAF.}(a). 
Most existing methods adopt a one-way guidance strategy of event-guided images, which is difficult to compensate for its structure.

To address this issue, we propose modeling modality guidance as a directional perceptual reconstruction mechanism, enabling complementary integration of heterogeneous features in cross-modal fusion. 
Specifically, structural and contextual information from the guiding modality is treated as prior knowledge and cues to selectively modulate, restore, and enhance the representation of the guided modality. 
To this end, we design the Bidirectional Guided Awareness Fusion (BGAF) module, which performs breakpoint-aware restoration and structure-aware enhancement through two successive stages of cross-modal interaction, as illustrated in \autoref{Figure5-BGAF.}(c) and (d). 
This module breaks the limitations of previous one-way fusion approaches by establishing tighter semantic coupling between the two modalities, significantly improving structural integrity and visual quality under low-light conditions.

\autoref{Figure3-BiLIE-network.}(c) illustrates the details of the BGAF module, which adopts a two-stage interactive guided attention mechanism to perform breakpoint-aware restoration and structure-aware enhancement for the two modalities at the feature level. 
In the first stage, the image modality guides the event modality. 
By leveraging the spatial continuity and structural consistency of the image, this stage effectively completes structural breakpoints and fills in missing color information within the event representation, thereby enhancing its overall completeness, as shown in \autoref{Figure5-BGAF.}(c).
In the implementation, we first apply $1\times 1$ convolution and $3\times 3$ depthwise separable convolution to the image feature $F_{I} \in R^{H\times W\times C}$ for channel mapping and local context modeling, generating the query  ($Q$). 
Meanwhile, the event feature $F_{E} \in R^{H\times W\times C}$ is projected into key ($K$) and value ($V$). 
Subsequently, the features are flattened and divided into multi-heads, and referring to the efficient attention mechanism proposed in \cite{shen2021efficient}, we construct an attention pathway with linear complexity, achieving efficient global cross-modal interaction through feature normalization and cross-modal matrix multiplication.
Finally, we add the cross-modal interaction features to the original image features via residual connections, implementing the introduction of structural priors from the event modality while maintaining the stability of the original image semantics.
The entire process can be represented as:
\begin{equation}
\begin{aligned} 
&CA_{1} \left ( F_{I},F_{E}   \right ) =\rho _{q} \left (Q  \right ) \left (\rho _{k} \left (  K\right ) ^{T}  V  \right ), \\
&Q=F_{I}W^{q}, K=F_{E} W^{k},V=F_{E}W^{v} ,  \\
&F_{I}^{'} =F_{I}+   CA_{1}\left ( F_{I} ,F_{E}  \right )  .
  \label{eq6}
\end{aligned}
\end{equation}
where $F_{I}^{'}$ represents the updated image features. 
$\rho _{q}$ and $\rho _{k}$ are the normalization functions for the query and key features respectively. 
$W^{q} ,W^{k} ,W^{v} \in C\times \left ( C/h \right )$ are three learnable parameter matrices, where $h$ is the number of heads in the multi-head cross-attention mechanism. 
Across our five levels, from low to high, $h$ is sequentially set as [2, 4, 4, 4, 6], enabling progressive feature alignment and semantic fusion from low resolution to high resolution.
This operation establishes local structural mapping relationships from images to events in the semantic space, effectively repairing the discontinuities in event features and compensating for the missing contextual structure.

In the second stage, the event modality guides the image modality. 
Leveraging the event camera’s high dynamic range and its differential sensitivity to luminance changes, this stage provides the image modality with clear contours and edges, as illustrated in \autoref{Figure5-BGAF.}(d). 
Specifically, the original event vector $F_{E} \in R^{H\times W\times C}$ is channel-mapped and modeled for local context to project into the query ($Q$), while the updated image vector $F_{I}^{'} \in R^{H\times W\times C}$ is projected to the key ($K$) and value ($V$). The attention computation is then performed in the same manner:
\begin{equation}
\begin{aligned} 
&CA_{2} \left ( F_{E} ,F_{I}^{'}  \right )  =\rho _{q} \left (Q  \right ) \left (\rho _{k} \left (  K\right ) ^{T}  V  \right ), \\
&Q=F_{E} W^{q} ,K=F_{I}^{'} W^{k} ,V=F_{I}^{'} W^{v}, \\
&F_{E}^{'}=F_{E} +  CA_{2} \left ( F_{E} ,F_{I}^{'}  \right )  . 
  \label{eq7}
\end{aligned}
\end{equation}
where $F_{E}^{'}$ represents the updated event features. Through this reverse guidance mechanism, the event features enhance the image modality's performance in edge and high-frequency texture areas, further improving the structural continuity and detail fidelity of the fused image.

Finally, we concatenate the guided enhanced features $F_{I}^{'}$ and $F_{E}^{'}$ output from both stages along the channel dimension and perform fusion mapping through a $1\times 1$ convolution to obtain the fused feature representation: 
\begin{equation}
  F_{f}^{i} = Conv\left ( Concat\left ( F_{I}^{'} ,F_{E}^{'}  \right )  \right )       
  \label{eq8}
\end{equation}
where $F_{f}^{i}$ is the output of the Bidirectional Guided Awareness Fusion (BGAF) module. 
This process is performed in parallel across all scales $i=1,2,\cdots ,5$ and involves upsampling and reconstructing layer by layer through a multi-scale decoder, ultimately producing an enhanced result with smooth textures and complete structure, as shown in \autoref{Figure5-BGAF.}(e). 
This achieves information complementarity and bidirectional enhancement between the two modalities.

\subsection{Loss Function}
Our total loss function is composed of four components:
\begin{equation}
  L_{total} =a\cdot L_{1} +  b\cdot L_{ML} +  c\cdot L_{FFT} +  d\cdot L_{colour} 
  \label{eq9}
\end{equation}
where $a,b,c,d$ are hyperparameters used to balance the four loss functions.
$L_{1},L_{ML},L_{FFT},L_{colour}$ represent $L_{1}$ loss, multi-level reconstruction loss, frequency loss, and color consistency loss, respectively.

\textbf{$L_{1}$ loss.} The pixel-level difference between the output of the model and the target is calculated at each scale:
\begin{equation}
  L_{1}=\sum_{l=1}^{N_{1} }  w_{l} \left \| f_{l}-y_{l}   \right \| _{1}   
  \label{eq10}
\end{equation}
where $N_{1}=5$ indicates 5 scales.
$w_{l}$ is the weight for the $l$-th layer.
$f_{l}$ and $y_{l}$ represent the model output and target output of the $l$-th layer.

\textbf{Multi-level reconstruction loss.} Following the multi-level reconstruction loss based on the variability of contrast distribution proposed in \cite{jiang2023event}, we generate images that are more similar to the ground truth while maintaining differences from the degraded images:
\begin{equation}
    L_{ML} =\sum_{l}^{N_{1} }\sum_{m}^{N_{2} }  \sum_{b}^{N_{3} } \frac{w_{l}\cdot \sigma _{m} \cdot lpips\left ( f_{l},y_{l}   \right )  }{lpips\left ( f_{l},I_{l}   \right ) }
  \label{eq11}
\end{equation}
where $lpips\left ( f_{l},y_{l}   \right )$ and $lpips\left ( f_{l},I_{l}   \right )$ represent the perceptual loss between the predicted output $f_{l}$ and the ground truth $y_{l}$ and the low-light image $I_{l}$ of the $l$-th layer, respectively. 
$H_{m}$ and $W_{m}$ represent the height and width of the feature map at the $m$-th layer.
$\mu _{m} $ is a set of learnable weight parameters.
$N_{2}=5$ indicates different feature layers.
$N_{3}=1$ indicates batch size.
$\sigma _{m}$ represents the weight of the similarity distance at the $m$-th layer.

\textbf{Frequency loss.} Low-light images and events often contain noise and artifacts, which may be amplified during the enhancement process.
Analyzing images in the frequency is a classic method for removing them.
Therefore, we use the loss function based on the Fast Fourier Transform (FFT) proposed in \cite{yadav2021frequency} to generate outputs with less noise:
\begin{equation}
  L_{FFT}^{\frac{H}{K}\times  \frac{W}{K} }\left (f,y  \right )  =\frac{K^{2}}{HW} \left | FFT\left ( f \right )  -FFT\left ( y \right ) \right | ^{\frac{H}{K}\times \frac{W}{K}  } 
  \label{eq12}
\end{equation}
where $f,y\in R^{H\times W\times C}$ are the predicted output and target image, with height and width denoted as $H$ and $W$. 
$K=\left [ 1,2,4,8,16 \right ]$ are scaling factors, indicating the calculation of frequency loss across five scales to obtain $L_{FFT}$.

\textbf{Color consistency loss.} We employ the color consistency loss function proposed in \cite{tang2023divfusion}, which constrains the color distribution using discrete cosine distance to align it more closely with the target image, thereby reducing color distortion introduced by the input image:
\begin{equation}
  L_{colour}=\frac{1}{HWC}\sum _{i\in \vartheta }   \sum_{n=1}^{N}cosine\left ( f,y \right ), \vartheta \in  \left \{ R,G,B \right \}   
  \label{eq13}
\end{equation}
where $i$ is an element in $\left \{ R,G,B \right \}$.
$N$ represents the number of pixels in the image.
$cosine\left ( f,y \right )$ indicates the cosine similarity between the output and the target image in the $i$-th channel.

\section{Experiments}
\label{sec:section 5}
\subsection{Datasets and Implementation Details}

\textbf{Datasets.} We evaluate our model on both the LIE \cite{jiang2023event} and the constructed RELIE dataset. 
The LIE contains 164 indoor and 42 outdoor scenes captured by the DAVIS346 camera. 
To facilitate a fair comparison of the visual quality between the two datasets, the training and testing splits of RELIE are kept identical to those of LIE in all experiments.

\textbf{Implementation Details.} We implement our proposed method in PyTorch and conduct training and testing on an NVIDIA GeForce RTX-4090. 
The batch size is set to 1. 
We use the Adam optimizer \cite{kingma2014adam} with an initial learning rate of 0.0005, which is adaptively reduced. 
During training and testing, all images are resized to 256×256.

\begin{table*}[ht]
\footnotesize
\centering
\setlength\tabcolsep{3pt}{
\begin{tabular*}{\linewidth}{cccccccc}
% \begin{tabularx}{\linewidth}{c c *{6}{>{\centering\arraybackslash}X}}
\toprule
\multirow{2}{*}{Input} & \multirow{2}{*}{Methods} & \multicolumn{3}{c}{RELIE} & \multicolumn{3}{c}{LIE} \\
& & PSNR & SSIM & LPIPS & PSNR & SSIM & LPIPS \\
\midrule
Event only & E2VID (TPAMI'19) & 14.10 & 0.354 & 0.542 & 9.08 & 0.259 & 0.579 \\
\midrule
& ZeroDCE (TPAMI'21) & 6.39 & 0.007 & 0.664 & 14.34 & 0.049 & 0.569 \\
Image only & SCI (CVPR'22) & 7.31 & 0.079 & 0.536 & 17.00 & 0.229 & 0.427 \\
unsupervised & NeRCo (ICCV'23) & 13.77 & 0.347 & 0.594 & 11.28 & 0.273 & 0.626 \\
& Zero-IG (CVPR'24) & 9.04 & 0.216 & 0.556 & 17.98 & 0.425 & 0.451 \\
\midrule
& SNR-Net (CVPR'22) & 16.77 & 0.595 & 0.445 & 23.39 & 0.723 & 0.371 \\
Image only & Retinexformer (ICCV'23) & 18.63 & 0.611 & 0.453 & 25.76 & 0.777 & 0.354 \\
supervised & RetinexMamba (ICONIP'24) & 16.60 & 0.544 & 0.479 & 24.88 & 0.758 & 0.360 \\
& Wave-Mamba (ACMMM'24) & 19.69 & \textcolor{blue}{0.697} & 0.576 & 25.89 & 0.863 & 0.332 \\
\midrule
& ELIE (TMM'23) & \textcolor{blue}{19.86} & \textcolor{red}{0.998} & \textcolor{blue}{0.365} & \textcolor{blue}{26.05} & \textcolor{blue}{0.878} & 0.270 \\
Event+Image & EvLight (CVPR'24) & 17.99 & 0.612 & 0.372 & 24.43 & 0.766 & \textcolor{blue}{0.264} \\
& BiLIE (Ours) & \textcolor{red}{20.67} & \textcolor{red}{0.998} & \textcolor{red}{0.342} & \textcolor{red}{26.33} & \textcolor{red}{0.999} & \textcolor{red}{0.258} \\
\bottomrule
\end{tabular*}
% \end{tabularx}
}
\caption{Quantitative comparisons on the RELIE and LIE datasets. The best and the suboptimal results are marked in \textcolor{red}{red} and \textcolor{blue}{blue}.}
\label{table 1}
\end{table*}

\textbf{State-of-the-Art Methods.} We compare our method with eleven advanced low-light enhancement methods (event-based E2VID \cite{rebecq2019high}, unsupervised image-based methods Zero-DCE \cite{li2021learning}, SCI \cite{ma2022toward}, NeRCo \cite{yang2023implicit}, and Zero-IG \cite{shi2024zero}, supervised image-based methods SNR-Net \cite{xu2022snr}, Retinexformer \cite{cai2023retinexformer}, RetinexMamba \cite{bai2024retinexmamba}, and Wave-Mamba \cite{zou2024wave}, and image-event fusion methods ELIE \cite{jiang2023event}, EvLight \cite{liang2024towards}). 
Among all methods, E2VID uses the pre-trained weights provided by the official source for testing, while the others are retrained on both datasets.

\textbf{Quantitative results.} We use Peak Signal-to-Noise Ratio (PSNR), Structural Similarity Index (SSIM) \cite{wang2004image}, and Learned Perceptual Image Patch Similarity (LPIPS) \cite{zhang2018unreasonable} as evaluation metrics. 
The quantitative results in \autoref{table 1} demonstrate that the event-based method (E2VID) performs better on RELIE than on LIE, but due to the lack of rich color information, its overall performance is still significantly behind.
Unsupervised frame-based methods, lacking direct constraints from reference ground truth, typically rely on the statistical properties or implicit priors of the data itself (such as brightness distribution and texture patterns) for enhancement, making it difficult to accurately learn the mapping between low-light images and ideal well-lit images, resulting in their performance generally being lower than supervised methods that can utilize ground truth for targeted optimization. 
In contrast, fusion methods that combine image and event information perform better under extreme low-light conditions. 
Specifically, our proposed BiLIE achieves state-of-the-art performance on both datasets.  
On the RELIE dataset, PSNR and LPIPS improve by 0.81dB and 0.023; on the LIE dataset, PSNR, SSIM, and LPIPS improve by 0.28dB, 0.121, and 0.006, respectively, fully validating the effectiveness of the proposed framework.

\subsection{Comparison with State-of-the-Arts}
\begin{figure*}
\centering 
\includegraphics[width=\linewidth, height=8cm]{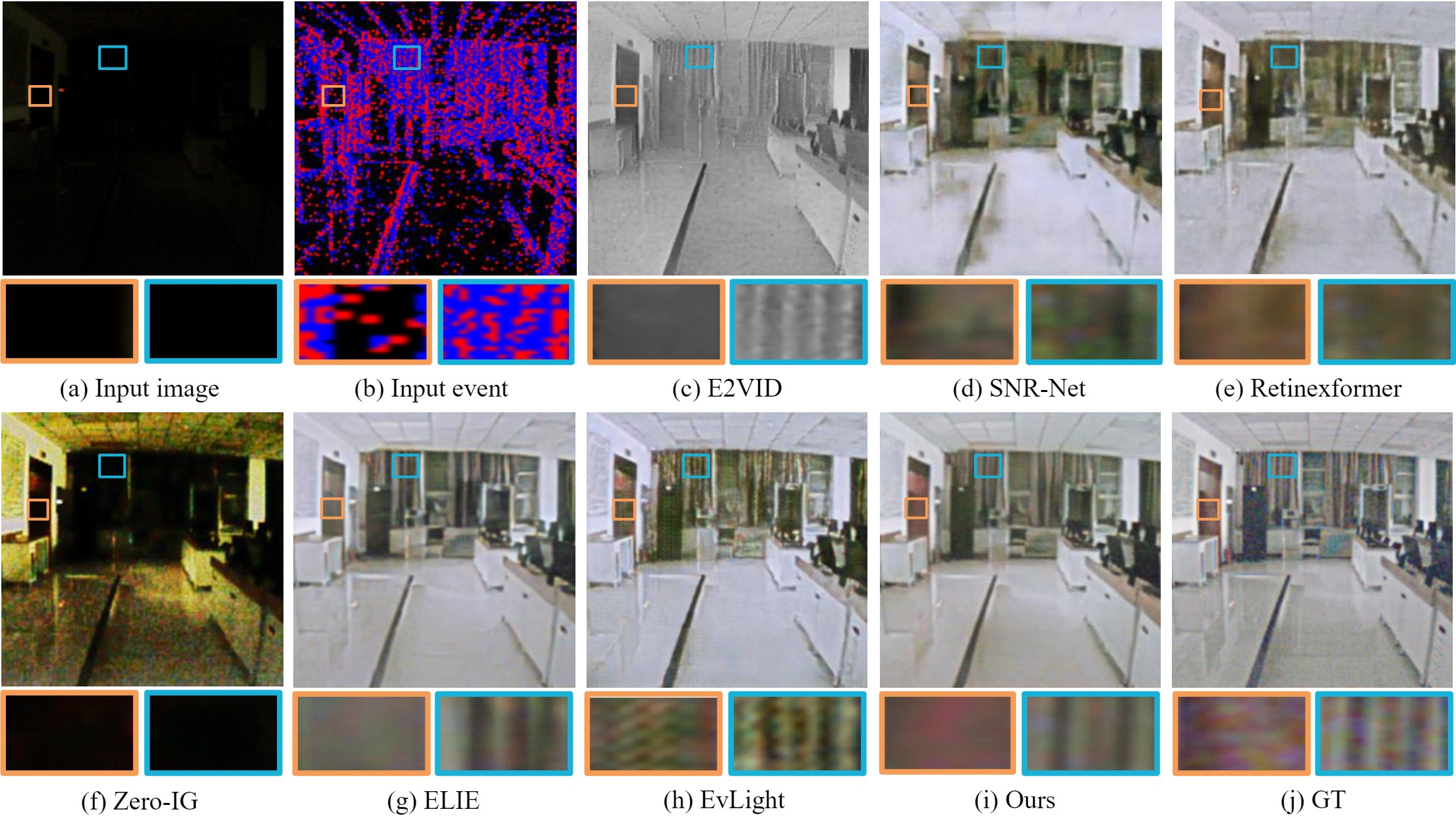} 
\caption{Qualitative results on indoor scenes from the RELIE dataset.} 
\label{Figure6-RELIE-indoor.} 
\end{figure*}

\begin{figure*}
\centering 
\includegraphics[width=\linewidth, height=8cm]{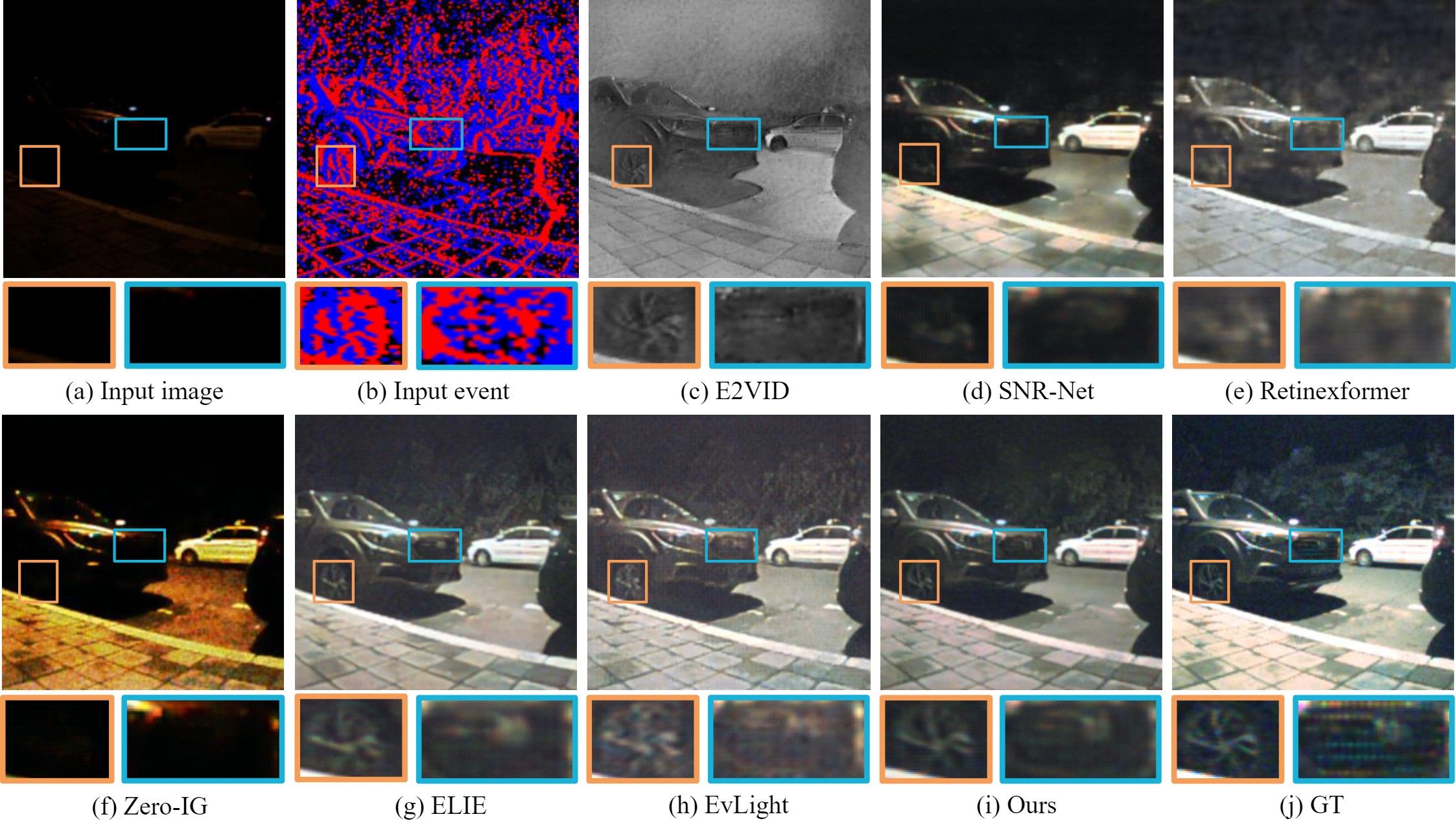} 
\caption{Qualitative results on outdoor scenes from the RELIE dataset.} 
\label{Figure7-RELIE-outdoor.} 
\end{figure*}

\textbf{Qualitative results.} \autoref{Figure6-RELIE-indoor.} and \autoref{Figure7-RELIE-outdoor.} present qualitative results on indoor and outdoor scenes from the RELIE dataset. 
Event-only methods such as E2VID suffer from poor visual quality due to the lack of color information, resulting in reconstructed images with low color diversity. 
Frame-only methods, including SNR-Net, Retinexformer, and Zero-IG, can restore brightness to some extent but often introduce noticeable artifacts and lose structural details.
In contrast, fusion-based methods that integrate both image and event modalities, such as ELIE, EvLight, and Ours, can better recover structural information in dark regions. 
However, ELIE exhibits color deviations (the red box in \autoref{Figure6-RELIE-indoor.}), while EvLight produces jagged noise and local discontinuities (the blue box in \autoref{Figure6-RELIE-indoor.}).
Our method outperforms existing approaches in both structural restoration and color consistency. 
It clearly reconstructs edges and details in dark regions while effectively suppressing local noise and light artifacts introduced during ground truth generation via unsupervised enhancement. 
The final output images exhibit higher naturalness, structural clarity, and visual coherence, with no significant artifacts or color shifts.

\begin{figure*}
\centering 
\includegraphics[width=\linewidth, height=8cm]{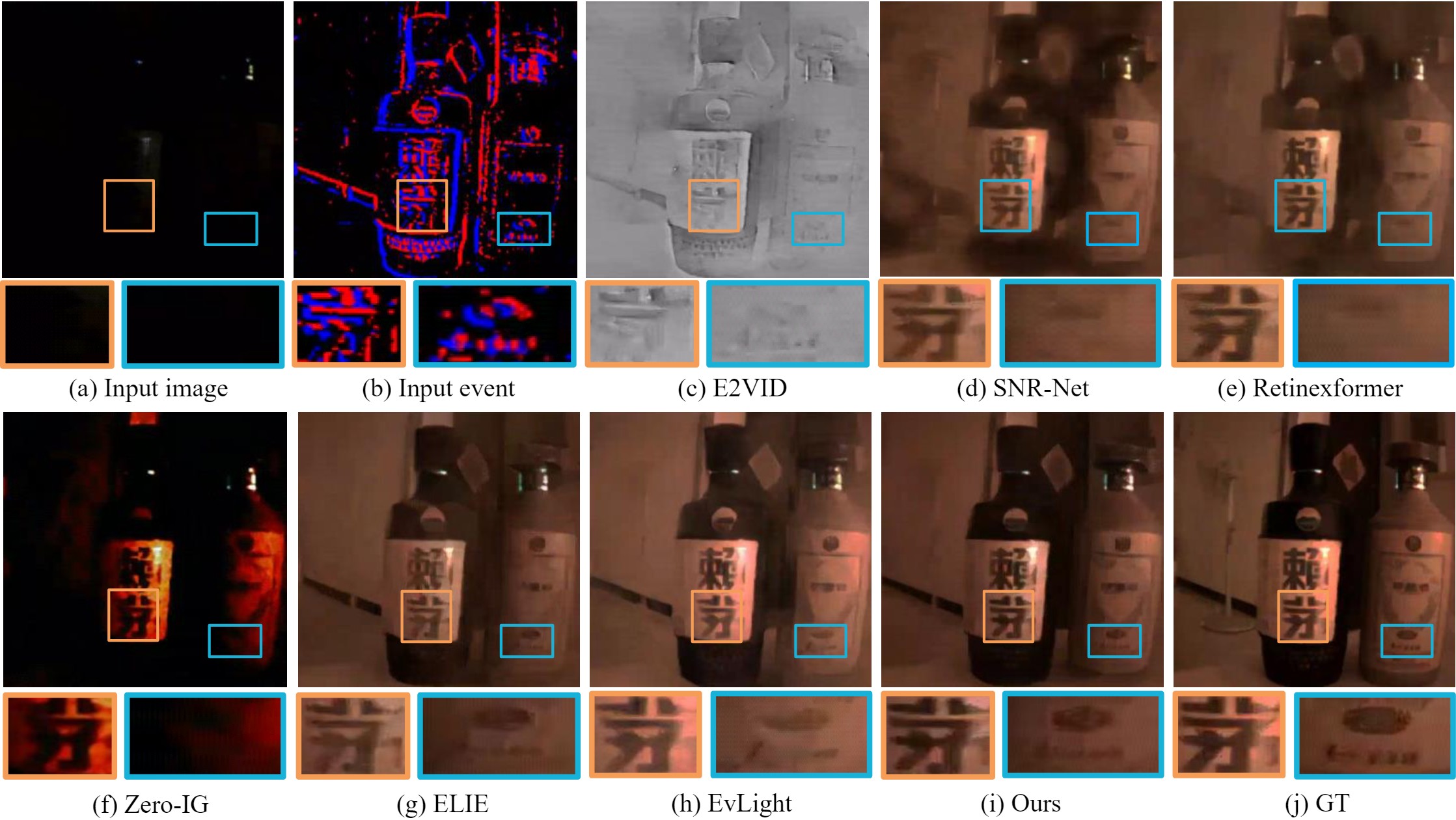} 
\caption{Qualitative results on indoor scenes from the LIE dataset.} 
\label{Figure8-LIE-indoor.} 
\end{figure*}

\begin{figure*}
\centering 
\includegraphics[width=\linewidth, height=8cm]{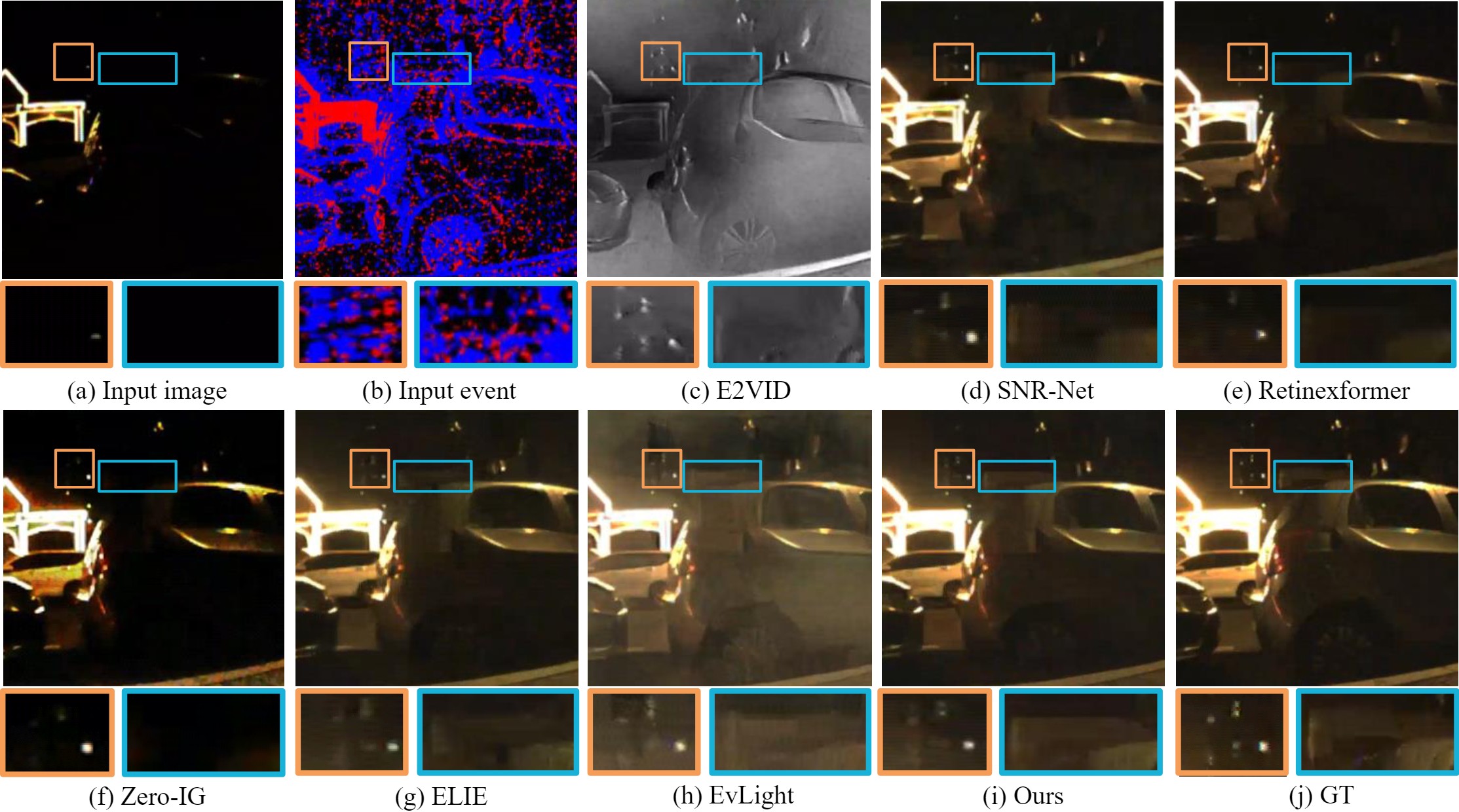} 
\caption{Qualitative results on outdoor scenes from the LIE dataset.} 
\label{Figure9-LIE-outdoor.} 
\end{figure*}

\autoref{Figure8-LIE-indoor.} and \autoref{Figure9-LIE-outdoor.} show qualitative results on indoor and outdoor scenes from the LIE dataset. 
Compared to the results on RELIE (\autoref{Figure6-RELIE-indoor.} and \autoref{Figure7-RELIE-outdoor.}), the original reference images in the LIE dataset suffer from lower contrast and color distortions, leading to noticeably inferior visual quality overall, manifested as color distortion and detail loss (the fan in \autoref{Figure8-LIE-indoor.}). 
This observation further demonstrates the advantage of our RELIE dataset in providing higher-quality ground truth and its positive impact on model training and performance.
Compared with frame-based methods (SNR-Net, Retinexformer, and Zero-IG), fusion-based methods (ELIE, EvLight, and Ours) exhibit more stable performance in challenging regions, validating the significance of introducing event information into the task of low-light image enhancement. Among these methods, ours achieves the best edge reconstruction and structural fidelity.

\subsection{Ablation Study}
We conduct a systematic ablation study on the RELIE dataset to evaluate the effectiveness of the key components in our proposed model. 
The results are summarized in \autoref{table 2}.
First, we adopt a naive baseline that simply concatenates \cite{han2020neuromorphic} image and event features without considering structural complementarity or noise suppression mechanisms across modalities, which yields the lowest enhancement performance (Row 1). 
When replacing this with the Bidirectional Guided Awareness Fusion (BGAF) module, the model effectively mitigates structural discontinuities caused by event sparsity and significantly enhances cross-modal structural alignment and information completion, resulting in a 0.06dB gain in PSNR (Row 2).
The Dynamic Adaptive Filtering Enhancement (DAFE) module adaptively suppresses global artifacts caused by sudden illumination changes and enhances high-frequency edge details in the event features, leading to sharper and clearer results, with a 0.14dB PSNR improvement (Row 3).
In addition, we perform ablation analysis on the frequency loss and color consistency loss. 
The results show that frequency loss helps remove noise and artifacts, providing stable performance gains (Rows 4 and 5). 
The color consistency loss alleviates color distortions introduced by chromatic bias in the input images and further improves overall model performance (Row 7).
In summary, the ablation results in \autoref{table 2} clearly demonstrate that each proposed module and loss function contributes positively to the model’s performance across different dimensions.

\begin{table}
\setlength\tabcolsep{5pt}
\centering
% \begin{tabular*}{\linewidth}{cccc ccc}
\begin{tabularx}{0.8\textwidth}{cccc ccc}
\toprule
\multicolumn{4}{c}{Methods} & \multicolumn{3}{c}{RELIE} \\
BGAF & DAFE & $L_{\mathrm{FFT}}$ & $L_{\mathrm{colour}}$ & PSNR $\uparrow$ & SSIM $\uparrow$ & LPIPS $\downarrow$ \\
\midrule
&  &   &  & 19.69 & 0.997 & 0.377 \\
\checkmark  &  &  &  & 19.75 & 0.998 & 0.366 \\
&  \checkmark  &  &  & 19.83 & 0.998 & 0.345 \\
\checkmark  &  & \checkmark  &  & 20.14 & 0.998 & 0.348 \\
&  \checkmark  & \checkmark  &  & 20.28 & 0.998 & 0.350 \\
\checkmark  & \checkmark  & \checkmark  &  & 20.22 & 0.998 & 0.354 \\
\checkmark  & \checkmark  & \checkmark  & \checkmark  & \textcolor{red}{20.67} & \textcolor{red}{0.998} & \textcolor{red}{0.342} \\
\bottomrule
% \end{tabular*}
\end{tabularx}
\caption{Ablation results of BiLIE on the RELIE dataset, where the model with all components achieves the highest performance, highlighted in \textcolor{red}{red}.}
\label{table 2}
\end{table}

In our proposed Dynamic Adaptive Filtering Enhancement (DAFE) module, a Gaussian high-pass filter is applied to the event features to suppress flickering artifacts introduced by global brightness fluctuations. 
The standard deviation ($\sigma$) of the Gaussian filter is a key parameter that controls its frequency response and has a significant impact on the filtering effect. 
A smaller $\sigma$ results in a sharper filter response, more effectively suppressing low-frequency components, but may also lead to excessive attenuation or loss of high-frequency information. 
In contrast, a larger $\sigma$ produces a wider frequency response, preserving more low-frequency content but potentially reducing image sharpness.
To balance the trade-off between low-frequency information suppression and high-frequency detail preservation, we conduct a systematic evaluation of the $\sigma_{1}$ parameter in the fixed filtering branch of the DAFE module on the RELIE dataset. 
As shown in \autoref{table 3}, we observe that the model achieves optimal performance when $\sigma_{1} = 12$, confirming the effectiveness and appropriateness of this parameter setting for the task.

In addition, to further evaluate the fusion strategy between the fixed filtering branch and the learnable filtering branch, we conduct a comparative analysis of different fusion methods, as shown in \autoref{table 3}. 
Compared with simple concatenation or direct addition, our proposed dynamic adaptive fusion mechanism adaptively adjusts the weight distribution between the two types of filtered features based on the input content, which achieves the highest PSNR by balancing stability and flexibility.

\begin{table}
\centering
\setlength{\tabcolsep}{3.8pt}{
\begin{tabular*}{\linewidth}{cccc|cccc}
% \begin{tabularx}{\linewidth}{c *{3}{>{\centering\arraybackslash}X}}
\toprule
Sigma & PSNR & SSIM & LPIPS & Fusion strategy &  PSNR & SSIM & LPIPS \\
\midrule
10 & 20.65 & 0.998 & 0.345 & Fixed filtering only & 20.09 & 0.998 & 0.347 \\
12 & \textcolor{red}{20.67} & \textcolor{red}{0.998} & \textcolor{red}{0.342} & Concatenation Fusion & 20.31 & 0.998 & \textcolor{red}{0.336} \\
14 & 20.67 & 0.998 & 0.346 & Addition Fusion & 20.47 & 0.998 & 0.346 \\
16 & 20.65 & 0.998 & 0.348 & Dynamic Fusion & \textcolor{red}{20.67} & \textcolor{red}{0.998} & 0.342 \\
\bottomrule
\end{tabular*}
% \end{tabularx}
}
\caption{Ablation study on the fixed filtering parameter and fusion strategy of DAFE.}
\label{table 3}
\end{table}

\begin{table}
\centering
\setlength{\tabcolsep}{8pt}{
\begin{tabular*}{0.8\textwidth}{cccc}
% \begin{tabularx}{\linewidth}{c *{3}{>{\centering\arraybackslash}X}}
\toprule
Structural configuration & PSNR $\uparrow$ & SSIM $\uparrow$ & LPIPS $\downarrow$ \\
\midrule
Image-guided Event & 19.20 & 0.998 & 0.445 \\
Event-guided image & 17.82 & 0.997 & 0.408 \\
Two-stage parallel & 20.55 & 0.998 & 0.343 \\
Two-stage sequential & \textcolor{red}{20.67} & \textcolor{red}{0.998} & \textcolor{red}{0.342} \\
\bottomrule
\end{tabular*}
% \end{tabularx}
}
\caption{Ablation study on structural configuration of BGAF.}
\label{table 4}
\end{table}

Finally, we conduct an in-depth ablation study on the architectural configurations of the Bidirectional Guided Awareness Fusion (BGAF) module. 
As shown in \autoref{table 4}, using only a single-stage guidance strategy, either image-to-event or event-to-image, significantly reduces model performance. 
This indicates that insufficient cross-modal interaction negatively impacts fusion quality.
We further compare a two-stage parallel concatenation structure with our proposed two-stage sequential design. 
The results demonstrate that the sequential configuration enables more effective step-by-step guidance and complementary enhancement, significantly outperforming the parallel strategy. 
This sequential modeling allows the image modality to first complete structural breakpoints in the event features, followed by event-guided refinement of the updated image features to enhance edges and dynamic details. 
The process exhibits stronger cross-modal interaction and structural restoration capabilities, ultimately achieving the best enhancement performance.

\subsection{Discussion and Analysis}
BiLIE achieves significant performance improvements in various experiments, mainly attributed to the following four core mechanisms: 
1) The Dynamic Adaptive Filtering Enhancement (DAFE) module achieves a dynamic balance between high-frequency enhancement and low-frequency suppression through dual-branch frequency domain filtering, effectively alleviating global flickering artifacts under rapidly changing lighting conditions while maintaining the clarity and structural stability of edge details.
2) The Bidirectional Guided Awareness Fusion (BGAF) module establishes bidirectional guidance relationships between image and event modalities through a two-stage guided attention mechanism. 
In the first stage, the image modality provides dense spatial priors to compensate for structural discontinuities in the event representation; 
in the second stage, the event modality offers dynamic information to reinforce texture and edge details, thus achieving better structural continuity and modal complementarity than unidirectional fusion.
3) The composite loss design combines frequency loss with color consistency loss, enhancing the naturalness and perceptual consistency of the enhancement results in terms of texture sharpness and color stability, making the generated images visually more authentic and stable. 
4) The RELIE dataset significantly improves the reliability and realism of the supervision signal through a systematic construction process and multidimensional evaluation scheme, providing the model training with a more stable optimization target and a more representative verification benchmark. 
Overall, BiLIE's overall performance advantage stems from the synergistic effect of each module and the joint promotion of data support.

Despite the considerable progress achieved in this study, certain limitations remain. 
First, the dual-branch attention interaction and multi-scale fusion structure increase computational complexity while enhancing performance, affecting the model's inference efficiency and real-time capability. 
Second, although the RELIE dataset improves in quality compared to existing datasets, it is still limited by the resolution and color bias of the DAVIS346 event camera and other hardware conditions, restricting the quality of data acquisition and the performance ceiling of the model. 
Moreover, the limited illumination and scene diversity in RELIE partially constrain the generalization capability and cross-scene adaptability of the model.

\section{Conclusion}
\label{sec:section 6}
This paper proposes an innovative low-light image enhancement framework (BiLIE), which incorporates bidirectional guidance between images and events. 
BiLIE effectively addresses two key challenges through its core components: the Dynamic Adaptive Filtering Enhancement (DAFE) module suppresses global artifacts in event features under dynamic lighting conditions, while the Bidirectional Guided Awareness Fusion (BGAF) module mitigates structural discontinuities caused by the spatial sparsity of event data.
In addition, we construct a high-quality, precisely aligned low-light image–event dataset (RELIE), which significantly improves the reference image quality of training samples and compensates for the lack of realism and usability in existing datasets.
Extensive experiments demonstrate that BiLIE achieves state-of-the-art performance on both the RELIE and LIE datasets, showing clear advantages in edge sharpness, noise suppression, and color fidelity.
In future work, we plan to adopt high-resolution event cameras and integrate them with RGB cameras to build a tri-axis cooperative multimodal hybrid imaging system for capturing higher-quality heterogeneous data. 
We will also explore more lightweight and efficient methods for low-light image enhancement to achieve higher performance and practical value.

\section*{Declaration of generative AI and AI-assisted technologies in the manuscript preparation process}
During the preparation of this work, the authors used Tencent HunYuan, Qwen2.5-VL-72B, and Gemini-2.5-Pro to assist in scoring each group of enhanced images during the construction of the RELIE dataset. These tools were applied to support the establishment of a multidimensional quality assessment mechanism, which was used to validate the superiority of the constructed dataset.
After using these tools, the authors carefully reviewed all outputs and take full responsibility for the content of the published article.

\section*{Acknowledgments}
This work was supported by the National Natural Science Foundation of China [Grant No. U24B20127]; the National Key Research and Development Program of China [Grant No. 2023YFC3081700]; and the Key Research and Development Program of Shaanxi Province [Grant No. 2023-LL-QY-24].

% \appendix
% \section{Example Appendix Section}


%%       \cite{<label>} ==> [1]

%%
% Example citation, See \cite{lamport94}.

 \bibliographystyle{elsarticle-num} 
 \bibliography{References}

\end{document}